\documentclass[runningheads]{llncs}
\usepackage{amsmath,amsfonts}
\usepackage{subcaption}
\usepackage{placeins}
\usepackage{hyperref}
\usepackage{graphicx}
\usepackage{caption}
\usepackage[T1]{fontenc}
\usepackage{graphicx,verbatim}
\begin{document}
\title{Neural Functional Alignment Space: Brain-Referenced Representation of Artificial Neural Networks}
\titlerunning{Neural Functional Alignment Space}
%

\author{Ruiyu Yan\inst{1} \and
Hanqi Jiang\inst{2} \and
Yi Pan\inst{2} \and
Xiaobo Li\inst{3} \and
Tianming Liu\inst{2} \and
Xi Jiang\inst{4} \and
Lin Zhao\inst{3}
}  
\authorrunning{R. Yan et al.}
\institute{
Tandon School of Engineering, New York University, Brooklyn, NY 11201, USA \\ 
\and
School of Computing, University of Georgia, Athens, GA 30602, USA\\
\and
Department of Biomedical Engineering, New Jersey Institute of Technology, Newark, NJ 07102, USA\\
\email{lin.zhao.1@njit.edu}
\and
The Clinical Hospital of Chengdu Brain Science Institute, MOE Key
Laboratory for NeuroInformation,School of Life Science and Technology, University of Electronic Science and Technology of China, Chengdu, 611731, China
}

\maketitle              
%



\begin{abstract}
We propose the Neural Functional Alignment Space (NFAS), a brain-referenced representational framework for characterizing artificial neural networks on equal functional grounds. NFAS departs from conventional alignment approaches that rely on layer-wise features or task-specific activations by modeling the intrinsic dynamical evolution of stimulus representations across network depth. Specifically, we model layer-wise embeddings as a depth-wise dynamical trajectory and apply Dynamic Mode Decomposition (DMD) to extract the stable mode. This representation is then projected into a biologically anchored coordinate system defined by distributed neural responses. We also introduce the Signal-to-Noise Consistency Index (SNCI) to quantify cross-model consistency at the modality level. Across 45 pretrained models spanning vision, audio, and language, NFAS reveals structured organization within this brain-referenced space, including modality-specific clustering and cross-modal convergence in integrative cortical systems. Our findings suggest that representation dynamics provide a principled basis for evaluating artificial intelligence in relation to biological computation.

\end{abstract}
\section{Introduction}
Understanding whether artificial intelligence recapitulates the computational principles of biological intelligence remains one of the most fundamental questions spanning neuroscience and machine learning. Although instantiated in distinct physical substrates, both brains and artificial neural networks transform sensory inputs into progressively abstract internal representations through hierarchies of nonlinear processing\cite{Yamins2014,Kell2018}. Recent empirical evidence has revealed striking correspondences between the representations in deep neural networks and neural activity measured in the human brain\cite{Schrimpf2020,Schrimpf2021,Goldstein2022}, suggesting that intelligent systems, regardless of implementation, may converge toward functionally analogous solutions under shared informational constraints\cite{Storrs2021,Huh2024}. These observations raise a deeper possibility: that intelligence may not be architecture-specific, but instead reflect an underlying computational organization that transcends physical embodiment~\cite{Hasson2020}.

Yet, as artificial models rapidly evolve from convolutional architectures to transformers and large-scale multimodal foundation models, the notion of “brain-AI alignment” remains inherently fragmented. Existing approaches typically quantify similarity at the level of layer-wise features or task-specific activations~\cite{Kriegeskorte2008,Schrimpf2020,Benchetrit2023}, yielding comparisons that are neither directly comparable across models nor invariant to modality or training objectives~\cite{Mehrer2020,Conwell2023}. Fundamentally, we lack a principled functional coordinate system in which heterogeneous intelligent systems can be evaluated on equal representational grounds. Notably, the human brain provides a naturally grounded reference frame: rather than comparing artificial models to one another, we may instead assess their intrinsic characteristics relative to distributed neural responses. From this perspective, brain–AI alignment ceases to be a qualitative correspondence and instead becomes a measurable property of intelligent information processing.

With a biologically anchored reference established, the remaining challenge is to identify a representation that captures a neural network’s intrinsic properties independent of architectural implementation. We argue that such a representation is not encoded in any single layer, but in the dynamical evolution of information as it propagates through network depth\cite{E2017,Chen2018}. In this view, stimulus representations trace structured trajectories in a latent representation space\cite{Cohen2020,Maheswaranathan2019}, and intelligence arises from the stability and consistency of these representational transformations rather than from isolated activation patterns\cite{Khona2024}.

Motivated by this perspective, we introduce the Neural Functional Alignment Space (NFAS), a brain-referenced framework that captures stable depth-wise network dynamics and embeds artificial models according to their functional similarity with distributed neural responses. Specifically, we model the evolution of layer-wise representations as a dynamical process and employ Dynamic Mode Decomposition (DMD) to extract a stable mode that characterizes stimulus-specific computation across depth~\cite{Schmid2010,Kutz2016}. The resulting representation is projected into a biologically anchored coordinate system via ROI-level encoding models\cite{Schaefer2018,Yeo2011}. We further define the Signal-to-Noise Consistency Index (SNCI) to quantify cross-model consistency while accounting for architectural variability. Evaluated on 45 models spanning vision, audio, and language using large-scale human fMRI benchmarks\cite{Cichy2021,LeBel2021,Li2022}, NFAS reveals structured organization across modalities and architectures, with integrative cortical systems exhibiting cross-modal convergence. Together, our framework provides a structured geometric scaffold that enables systematic positioning and comparative analysis of heterogeneous neural networks within a common brain-defined space.

\section{Related Works}
\textbf{Brain-DNN Alignment.} Recent studies have revealed correspondences and alignments between deep neural networks and cortical processing stages across vision~\cite{Cohen2020,Yamins2014}, audition~\cite{Kell2018}, and language~\cite{Goldstein2022,Schrimpf2021,Antonello2024}. These findings suggest that deep neural networks can capture modality-specific representation structure observed in biological neural systems~\cite{Schrimpf2020}. However, most alignment studies remain confined to individual modalities, limiting cross-modal comparison and generalization.
Here, we define a Neural Functional Alignment Space that enables architecture- and modality-agnostic comparison of neural networks based on their brain alignment profiles.


\textbf{Hierarchical Feature Modeling.} Existing alignment methods typically collapse network depth into layer selection, either by selecting peak-correlated layers~\cite{Schrimpf2020,Yamins2014} or identifying shared subspaces~\cite{Raghu2017,Whittington2020}. Such approaches treat layers as independent units, implicitly assuming that a single layer can sufficiently represent the model as a whole, thereby overlooking transformations induced by depth. Sequence-aware methods analyze representational trajectories~\cite{Chen2018,Maheswaranathan2019}, but depth modeling and feature aggregation remain modality-specific, limiting cross-domain comparability~\cite{Mehrer2020}. Our formulation instead interprets depth as an ordered representational evolution summarized within a shared analytical space.

\textbf{Correlation-based Brain Scores.}  Neural alignment is typically quantified using correlation-based metrics averaged across models within a modality~\cite{Kriegeskorte2008,Schrimpf2020}. While suitable for single-modality evaluation, simple averaging does not explicitly capture cross-model variability and limits comparability across heterogeneous domains~\cite{Conwell2023,Mehrer2020,Storrs2021}. Therefore, we introduce the Signal-to-Noise Consistency Index (SNCI), which integrates within-modality variability to attenuate architecture-specific outliers and identify alignment patterns consistently shared across models of the same modality.

\section{Method}
\subsection{Overview}
\begin{figure}[htbp]
\centering
\includegraphics[width=1.0\linewidth]{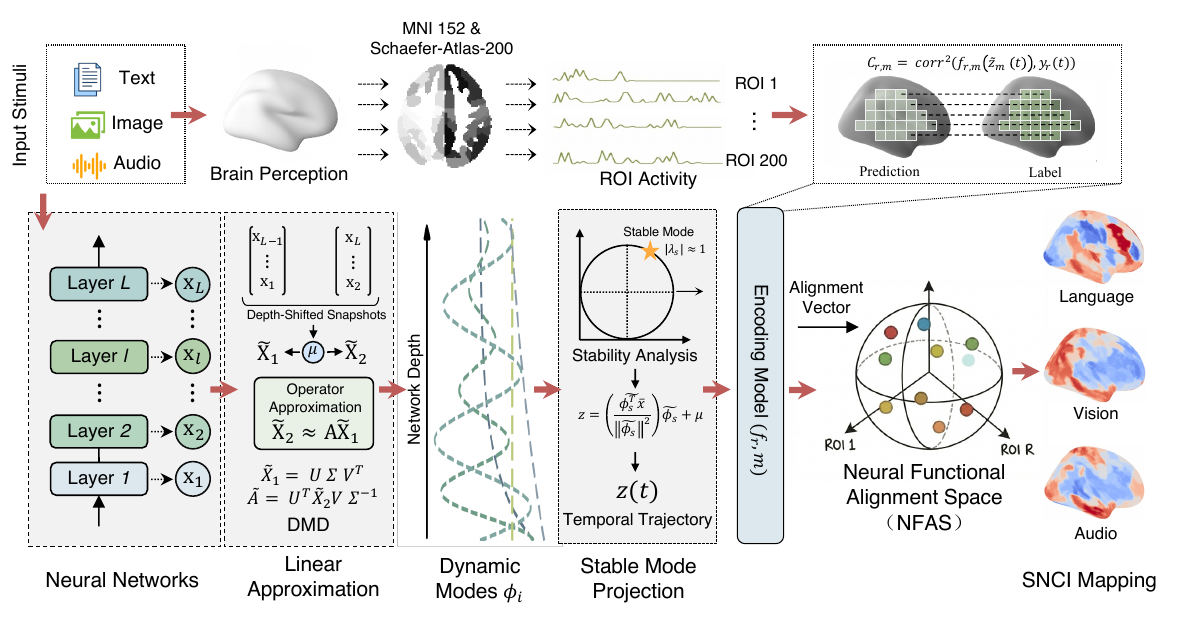}
\caption{Overview of the Neural Functional Alignment Space Framework}
\label{fig:fig1}
\end{figure}

Our framework is illustrated in Fig.~\ref{fig:fig1}. We first model layer-wise embeddings as depth-wise network dynamics to extract stable stimulus representations (Section~\ref{sec:NNDM}), then align these representations to fMRI neural responses to construct a Neural Functional Alignment Space (NFAS) for brain-referenced model comparison (Section~\ref{sec:NFAS}), and finally introduce the Signal-to-Noise Consistency Index (SNCI) for robust modality-level analysis (Section~\ref{sec:SNR}).

\subsection{Neural Network Dynamics Modeling}
\label{sec:NNDM}
Given an input (text, image, or audio), a deep neural network produces feature embeddings across layers. We model layer-wise embeddings as undergoing depth-wise dynamical evolution rather than as independent components.

Formally, let $\mathbf{x}_\ell \in \mathbb{R}^{D}$ denote the feature embedding at the $\ell$-th layer of a network with $L$ layers, where $\ell = 1,\dots,L$. The ordered sequence $\{\mathbf{x}_\ell\}_{\ell=1}^{L}$ characterizes the depth-wise dynamical evolution of the input representation. To characterize the dynamics governing this evolution, we construct two depth-shifted snapshot matrices:
\begin{equation}
\mathbf{X}_1 =
\begin{bmatrix}
\mathbf{x}_1, \mathbf{x}_2, \cdots, \mathbf{x}_{L-1}
\end{bmatrix}
\in \mathbb{R}^{D \times (L-1)},
\qquad
\mathbf{X}_2 =
\begin{bmatrix}
\mathbf{x}_2, \mathbf{x}_3, \cdots, \mathbf{x}_{L}
\end{bmatrix}
\in \mathbb{R}^{D \times (L-1)}.
\end{equation}

Although deep neural networks are non-linear, architectures such as ResNet~\cite{He2016} and Transformer~\cite{Vaswani2017} employ residual connections that facilitate smooth feature transitions across layers, making the evolution mathematically akin to discretized ordinary differential equations~\cite{Chen2018,E2017}. Motivated by Koopman operator theory~\cite{Lindner2024}, we approximate this evolution with a linear operator:
\begin{equation}
\mathbf{X}_2 \approx \mathbf{A}\mathbf{X}_1,
\end{equation}

To ensure compatibility with this linear evolution model, we first remove global offsets by subtracting the cross-layer mean, yielding centered snapshots:
\begin{equation}
\boldsymbol{\mu}
=
\frac{1}{L-1}
\sum_{\ell=1}^{L-1}
\mathbf{x}_\ell
\in \mathbb{R}^{D},
\quad
\tilde{\mathbf{X}}_1 = \mathbf{X}_1 - \boldsymbol{\mu}\mathbf{1}^\top,
\quad
\tilde{\mathbf{X}}_2 = \mathbf{X}_2 - \boldsymbol{\mu}\mathbf{1}^\top,
\end{equation}
where $\mathbf{1} \in \mathbb{R}^{(L-1)\times 1}$ denotes an all-ones vector. We then estimate the operator $\mathbf{A}$ via DMD~\cite{Schmid2010}. Specifically, we compute the singular value decomposition and derive the reduced-order operator:
\begin{equation}
\tilde{\mathbf{X}}_1 = \mathbf{U}\boldsymbol{\Sigma}\mathbf{V}^\top,
\quad
\tilde{\mathbf{A}} = \mathbf{U}^\top \tilde{\mathbf{X}}_2 \mathbf{V}\boldsymbol{\Sigma}^{-1}.
\end{equation}

Eigen-decomposition of $\tilde{\mathbf{A}}$ yields eigenvalues $\{\lambda_i\}$ and DMD modes $\{\boldsymbol{\phi}_i\}$, which characterize distinct patterns of depth-wise evolution. For dynamic mode $\boldsymbol{\phi}_i$, its evolution is governed by the associated eigenvalue $\lambda_i$~\cite{Khona2024,Kutz2016}, where $|\lambda_i|$ indicates amplification ($>1$), decay ($<1$), or stability ($\approx 1$)~\cite{Cohen2020,Huh2024}. We therefore select the mode $\tilde{\boldsymbol{\phi}}_s$ whose eigenvalue magnitude is closest to unity as the most stable mode. We then project the depth-averaged representation onto the selected dynamical mode and add back the original mean to recover the absolute baseline, yielding the stable dynamical representation $\boldsymbol{z}$ for a single stimulus:

\begin{equation}
\bar{\boldsymbol{x}} =
\frac{1}{L}
\sum_{\ell=1}^{L}
\boldsymbol{x}_\ell
\in \mathbb{R}^{D},
\quad
\boldsymbol{z}
=
\left(
\frac{\tilde{\boldsymbol{\phi}}_s^\top \bar{\boldsymbol{x}}}
{\|\tilde{\boldsymbol{\phi}}_s\|^2}
\right)
\tilde{\boldsymbol{\phi}}_s
+ \boldsymbol{\mu}.
\end{equation}

When stimuli are presented continuously over time, the model produces a corresponding sequence of representations, forming a temporal evolution $\boldsymbol{z}(t)$.

\subsection{Neural Functional Alignment Space}
\label{sec:NFAS}
To align network dynamics with fMRI responses, we convolve the resulting time series $\boldsymbol{z}(t)$ with a canonical hemodynamic response function (HRF)\cite{Friston1998}, yielding $\tilde{\mathbf{z}}(t)$. We define a Neural Functional Alignment Space (NFAS), a biologically anchored embedding space in which each axis corresponds to a functional ROI. Neural networks can be projected into this space according to its representational similarity with each ROI, yielding a brain-referenced network embedding.

Specifically, for each ROI $r$, we fit a linear encoding model $f_{r,m}$ mapping the $\tilde{\mathbf{z}}_m(t)$ of neural network $m$ to the corresponding ROI response $y_r(t)$. Alignment is quantified by the squared Pearson correlation between predicted and observed neural responses, yielding an ROI-level alignment score $c_{r,m}$ for model $m$:

\begin{equation}
c_{r,m}
=
\mathrm{corr}^2
\big(
f_{r,m}(\tilde{\mathbf{z}}_m(t)),
\, y_r(t)
\big),
\quad
\mathbf{c}_m
=
\left( c_{1,m}, \dots, c_{r,m} \right)
\in \mathbb{R}^R .
\end{equation}

For each model, the collection of scores across ROIs forms an alignment vector $\mathbf{c}_m$, and the set of such vectors spans the $R$-dimensional NFAS. Within this space, neural networks are represented as points according to their alignment profiles, enabling direct comparison based on their representational similarity to distributed brain regions. NFAS provides a brain-referenced coordinate system for systematically comparing similarity and variability across neural networks.

\subsection{Signal-to-Noise Consistency Index (SNCI)}
\label{sec:SNR}
While NFAS enables geometric comparison of individual networks, extending it to modality-level analysis is non-trivial. A naïve approach averages alignment vectors within each modality, but heterogeneous architectures can bias such aggregation through extreme models and obscure inter-model variability. We therefore introduce the Signal-to-Noise Consistency Index (SNCI), which quantifies alignment relative to cross-model variability within each modality~\cite{Nili2014}.

For each ROI $r$, we define the mean alignment across models within a given modality and its variability as:
\begin{equation}
\mu_r = \frac{1}{M} \sum_{m=1}^{M} c_{r,m},
\qquad
\sigma_r = \sqrt{ \frac{1}{M} \sum_{m=1}^{M} \left( c_{r,m} - \mu_r \right)^2 }.
\end{equation}

To enable comparison, SNCI is bounded in $[0,1]$ via a sigmoid transformation:
\begin{equation}
\mathrm{SNCI}_r
=
\mathrm{sigmoid}\!\left( \frac{\mu_r}{\sigma_r + \epsilon} \right)
\end{equation}
where $\epsilon$ is a small constant for numerical stability.

High SNCI values indicate strong and cross-architecture consistent alignment within a modality, whereas low values reflect weak or variability-driven correspondence. SNCI therefore highlights alignment patterns that generalize across architectures rather than those arising from isolated high-magnitude models.

\section{Experiments}
\subsection{Datasets and Implementation Details}
We evaluate our framework on three large-scale public fMRI benchmarks spanning audio, vision, and language: Narratives~\cite{LeBel2021}, Algonauts 2021~\cite{Cichy2021}, and The Little Prince~\cite{Li2022}. Brain data were preprocessed following benchmark protocols, normalized to MNI152 space~\cite{Fonov2011}, and parcellated with the Schaefer-200 atlas~\cite{Schaefer2018}. ROI time series were computed by averaging voxel-level BOLD signals within each parcel.

To assess cross-architecture stability, we analyze 45 frozen pretrained models covering diverse architectures and scales across modalities. The audio set includes Wav2Vec 2.0 and related self-supervised speech models~\cite{Hsu2021,Baevski2022}; the vision set spans multiple transformer-based architectures~\cite{Dosovitskiy2021,Oquab2024}; and the language set comprises BERT-family and cross-lingual transformers~\cite{Devlin2019}.

Layer-wise representations are fused to obtain the stable representation $\boldsymbol{z}(t)$, convolved with the hemodynamic response function (HRF) to yield $\tilde{\mathbf{z}}(t)$~\cite{Friston1998}, and alignment is quantified as squared Pearson correlation $c_{r,m}$ for each ROI $r$ and model $m$, yielding ROI-level vectors $\mathbf{c}_m$. The collection $\{\mathbf{c}_m\}$ across models defines the NFAS, while modality-level stability is summarized using $\mathrm{SNCI}_r$.

\subsection{Geometric Structure of the Neural Functional Alignment Space}
\begin{figure}[htbp]
    \centering
    \includegraphics[width=0.9\linewidth]{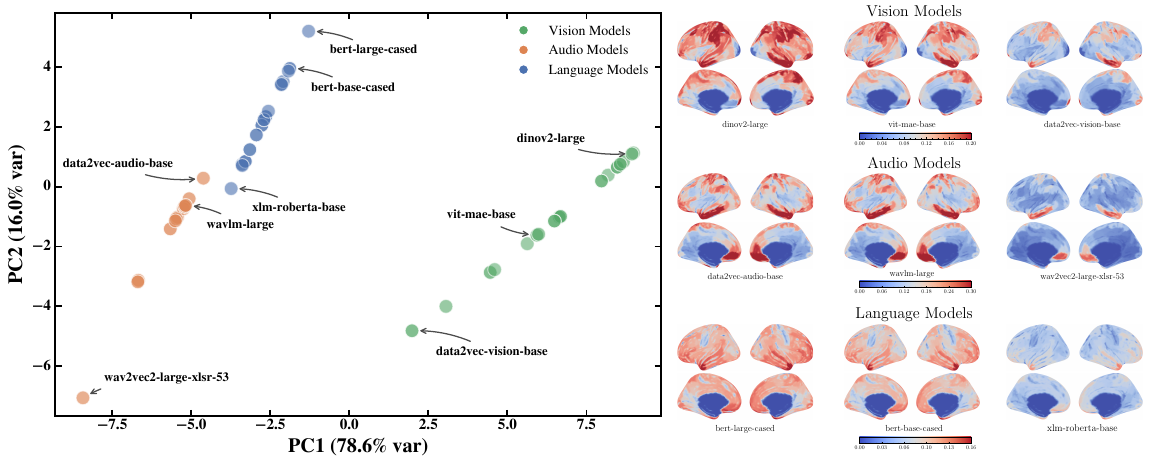}
    \caption{Structured Geometry of the Neural Functional Alignment Space.}
    \label{fig:fig2}
\end{figure}

The collection of alignment vectors $\{\mathbf{c}_m\}$ across models defines the Neural Functional Alignment Space. We visualized the NFAS at two complementary levels (Fig.~\ref{fig:fig2}): projecting the alignment vectors into a low-dimensional space using principal component analysis (PCA)~\cite{Jolliffe2002}, and mapping each alignment onto the cortical surface to reveal its ROI-level alignment topology.

In the reduced embedding, models cluster clearly by modality, indicating systematic differences in cortical alignment profiles. This geometric separation was statistically confirmed using PERMANOVA with 999 permutations, which revealed a significant effect of modality on NFAS structure ($p < 0.001$). Inter-modality distances substantially exceeded intra-modality distances (mean cosine distance $0.060$ vs.\ $0.0004$)\cite{Anderson2001}, and silhouette analysis indicated strong clustering strength ($S = 0.995$, permutation $p < 0.001$), indicating robust modality-dependent organization in the alignment space.

These results demonstrate that the NFAS captures geometric organization across models: modalities are clearly separable, while models within each modality share common alignment structure yet retain architecture-specific variation.

\subsection{Cortical Organization of Modality-Level Alignment}
\begin{figure}[htbp]
    \centering
    \includegraphics[width=0.95\linewidth]{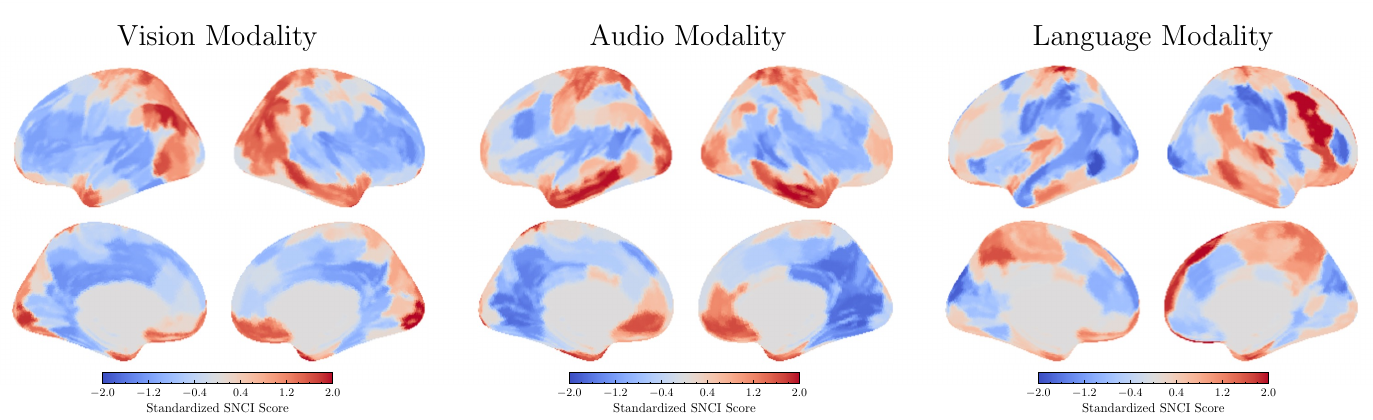}
    \caption{Modality-Specific Cortical Distribution of SNCI.}
    \label{fig:fig3}
\end{figure}

To examine cortical organization, we projected ROI-level SNCI scores onto the cortical surface (Fig.~\ref{fig:fig3}), using z-score standardization across ROIs for visualization. The maps reveal coherent large-scale alignment topography rather than diffuse patterns, consistent with established cortical functional organization.

Visual models show strong posterior alignment in occipital cortex, particularly early visual areas such as V6 and V7. Audio models exhibit dominant alignment in lateral temporal regions associated with higher-level auditory processing. Language models show peak alignment in prefrontal control regions, consistent with executive and semantic integration systems.

These spatial patterns indicate that each modality preferentially aligns with cortical systems specialized for corresponding human sensory or cognitive functions while preserving structured large-scale organization.

\subsection{Functional Network-Level Comparison}

\begin{figure}[htbp]
    \centering
    \includegraphics[width=0.85\linewidth]{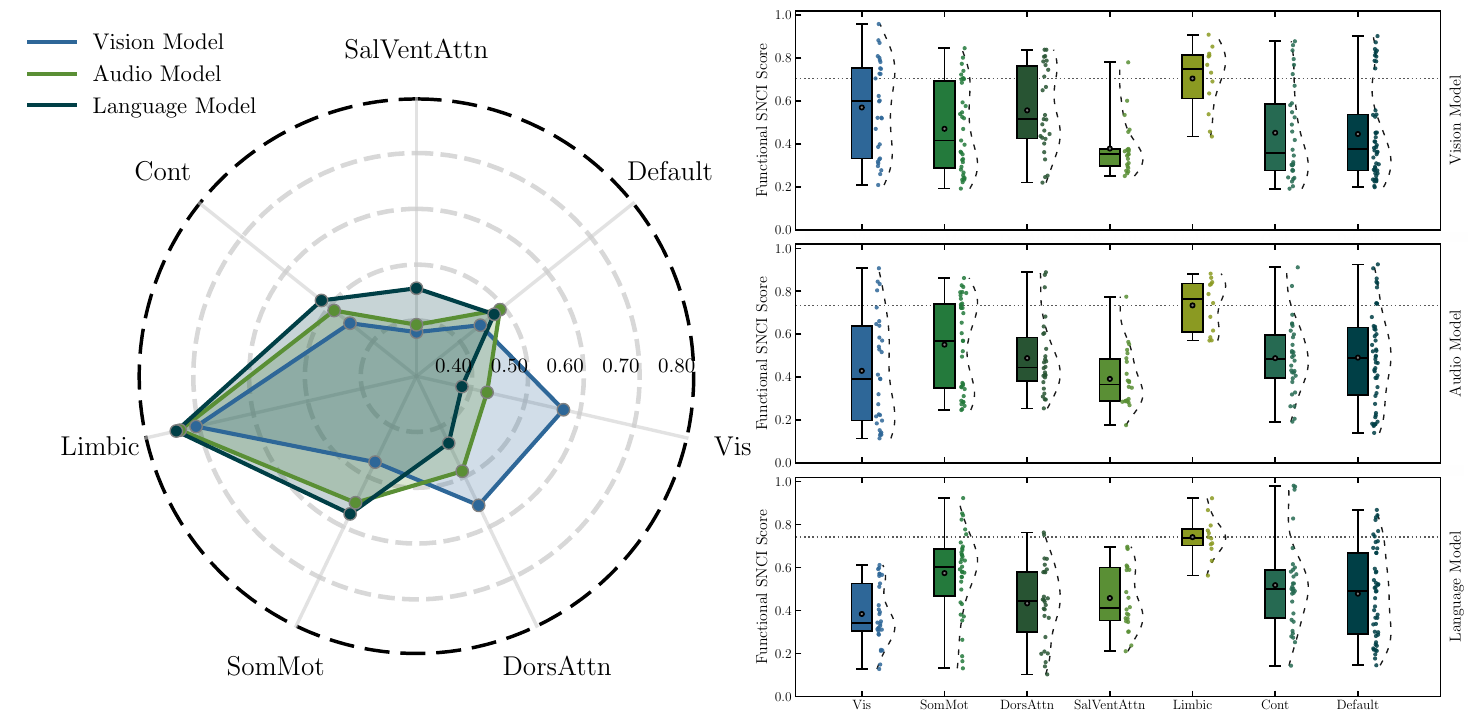}
    \caption{Comparison of brain SNCI across functional networks.}
    \label{fig:fig4}
\end{figure}

We further aggregated ROI-level SNCI scores according to the Yeo-7 functional networks and averaged within each system (Fig.~\ref{fig:fig4})~\cite{Yeo2011}. Radar plots and boxplots were used to assess within-system consistency and cross-model variability. A two-way ANOVA with factors modality and network revealed significant main effects and a significant modality $\times$ network interaction (all $p < 0.001$), demonstrating that alignment differences reflect network-specific patterns rather than uniform magnitude shifts across modalities.

Across modalities, the Limbic network shows relatively elevated alignment. Given its role in emotion, memory, and contextual integration, this pattern suggests convergence with affective and memory-related systems. Beyond this shared component, modality-specific specialization emerges: visual models show additional elevation in the Visual network, audio models in attention-related systems, and language models broader alignment across association networks, particularly Default and Control systems.

Overall, modality-specific models exhibit structured organization paralleling large-scale cortical systems engaged in related sensory and cognitive processes.

\section{Discussion}
This study reframes Brain–Neural Network alignment as geometric organization within a brain-defined space. By modeling network depth as an ordered representational evolution rather than discrete layers, alignment emerges as a structured property of representational trajectories. The resulting Neural Functional Alignment Space (NFAS) reveals intrinsic geometric organization in model–brain correspondence across modalities and architectures.

Within NFAS, heterogeneous models are compared on equal functional grounds through distributed ROI-level alignment profiles. Modality-level clustering suggests that architectural differences manifest as structured displacements within a shared cortical embedding, while convergence in integrative networks indicates shared computational constraints beyond modality-specific processing.

By incorporating SNCI as a variability-aware statistic, the framework highlights cross-architecture consistency. More broadly, it situates Brain–AI comparison within a topology-aware paradigm, providing a principled basis for examining how artificial systems relate to large-scale brain organization.

    

%
%
%
%

\end{document}